**THIS IS A PREPRINT. IT HAS NOT BEEN PEER-REVIEWED**

# Using role-play and Hierarchical Task Analysis for designing human-robot interaction

Mattias Wingren[1] [0000-0003-4269-5093], Sören Andersson[1] [0000-0003-0418-185X], Sara Rosenberg[2 3] [0000-0002-8938-7939], Malin Andtfolk[3] [0000-0001-9380-3619], Susanne Hägglund[1 3] [0000-0002-4941-6811], Prashani Jayasingha Arachchige[4] [0000-0003-4168-9732], and Linda Nyholm[3 5] [0000-0002-3324-5966]

[1] Experience Lab, Faculty of Education and Welfare studies, Åbo Akademi University, Rantakatu 2, 65100 Vaasa, Finland
[2] Pharmaceutical Sciences Laboratory, Faculty of Science and Engineering, Åbo Akademi University, Tykistökatu 6A, 20520 Turku, Finland
[3] Department of Caring Science, Faculty of Education and Welfare studies, Åbo Akademi University, Rantakatu 2, 65100 Vaasa, Finland
[4] Department of Information Technology, Faculty of Science and Engineering, Åbo Akademi University, Vesilinnantie 3, 20500 Turku, Finland
{mattias.wingren, soren.andersson, sara.rosenberg, malin.andtfolk, susanne.hagglund, prashani.jayasinghaarachchige}@abo.fi
[5] Department of Caring and Ethics, Faculty of Health Sciences, University of Stavanger, Kjell Arholms gate 43, Stavanger, Norway
linda.s.nyholm@uis.no

**Abstract.** We present the use of two methods we believe warrant more use than they currently have in the field of human-robot interaction: role-play and Hierarchical Task Analysis. Some of its potential is showcased through our use of them in an ongoing research project which entails developing a robot application meant to assist at a community pharmacy. The two methods have provided us with several advantages. The role-playing provided a controlled and adjustable environment for understanding the customers' needs where pharmacists could act as models for the robot's behavior; and the Hierarchical Task Analysis ensured the behavior displayed was modelled correctly and aided development through facilitating co-design. Future research could focus on developing task analysis methods especially suited for social robot interaction.

## 1    Introduction

The design process of robot applications can incorporate a vast number of methods, being the interdisciplinary field that it is. Sources for new inspiration have often been the fields of human-computer interaction, interaction design and user experience design. A clear example of this influence can be seen in the design of robot applications according to the principles of user-centered design, e.g., [1, 2, 3]. What have not been used as extensively, however, is role-play [4] and task analysis [5], especially when the robots are social robots. The aim of this paper is to bring attention to these, in our opinion, neglected methods by showing their usefulness in the field of human-robot interaction (HRI).

We will do this in the following fashion: first, we will present some background to the two methods and the role they have played in the fields from where we draw inspiration. Secondly, we will showcase their use in our ongoing research project where we are designing a robot application on the Furhat platform; the application is meant to offer medication counselling regarding emergency contraceptive pills to community pharmacy customers [6]. The pharmaceutical context is an area where social robots have seen little use, and we hope an introduction of our application can — among other things — offload pharmacists' high workload and provide non-judgmental counselling in the sensitive situation that buying an emergency contraceptive pill is.

## 2    Background

### 2.1    Simulation and role-play

Simulation has a rich history in human factors, the study and design of human-machine systems [7]. One handbook provides the following definition of simulation: "A method for implementing a model over time. Also, a technique for testing analysis or training in which real-world systems are used or where real-world and conceptual systems are reproduced by a model" [8]. Many kinds of simulations fit into this broad definition, a few examples being computer simulations [9], war game simulations [10], aviation simulations [11] and role-playing [12] — all which substitute phenomena with simpler ones. This last type of simulation, which can be seen as people acting according to a given role [4], has seen a fair bit of use in neighboring design fields like interaction design and user experience design [13, 14, 15].

Human-robot interaction is no stranger to simulation either, but mainly presents itself in two different forms. (1) Either as computer simulations where a robot is simulated for easier manipulation and programming [16], or (2) role-playing where one person controls a robot but pretends it moves on its own volition, an often-used method called Wizard of Oz [17]. There are, however, exceptions. One study, for example, designed a system which translated bodystormed ideas to editable robot applications [18]; another, role-played hazardous scenarios involving robots to better prepare for them in the future [19]; and a third role-played interactions with a robot in the design phase for further ideation [20]

A promising, but in our opinion underutilized, methodology is using role-playing as a kind of blueprint in the design process [21]. If your goal, for example, is to have a robot behave as a playing partner in a game, you can first observe a human exhibiting this behavior and then model the robot's behavior upon this [22]. This is a lot easier than starting from scratch when designing the interaction, especially if the designers are not knowledgeable about the role the robot is meant to fulfil (for example non-teachers designing a robot tutor). Role-playing also has the advantage of being a more adjustable and controlled environment [23]. It enables creative experimentation not easily obtained through traditional means. If, for example, a study requires an older person to display anger and contempt towards a robot to test the latter's usability, this is much more feasibly done by having someone play this specific role instead of trying to organically stage this occurrence. Role-play can also minimize the risk a real environment poses [24]. A social interaction in a hospital, for example, may be problematic to study since it breaches on privacy and in some cases might even present danger. In cases like these, the environment can be simulated to circumvent these kinds of hazards. Finally, role-playing also adds to the more traditional way of conducting user research, such as interviewing and observing [25], a crucial step for involving and empowering the end-users. Interviews are a powerful tool when done right, but a lot of a persons' skill and knowledge cannot easily be articulated [26], and even that which can be articulated might in some cases be post-hoc constructions [27]. Traditional observation, on the other hand, is a powerful approach, but it is not a given the observee will exhibit the behaviors the observer is interested in. When role-playing, however, the observers can, as mentioned before, stage the wanted scenarios by controlling the environment.

### 2.2  Task analysis and Hierarchical Task Analysis

A professional's seemingly easy task often belies greater complexity where they juggle between actions based on intuition, skills and rules [28]. Just like simulation, human factors have long been interested in modelling the tasks of professionals, and the common name for methods used to gain understanding of a task is, as the name implies, task analysis [29]. Arguably the most popular of these is Hierarchical Task Analysis (HTA), which divides the task into smaller goals and sub-goals [30]. If, for example, one was to analyze the task of hammering a nail [31], the overreaching goal could be hammering the nail into the plank (0), the subgoals steadying the nail (1) and hammering (2), whereas a sub-goal of hammering could be raising (2.1) and striking the hammer (2.2) (see figure 1). One also wants rules directing the order and behavior of the goals, for example starting with steadying the nail before hammering (plan 0) and stopping when the nail is flush with the plank (plan 2); this is implemented in the way of plans.

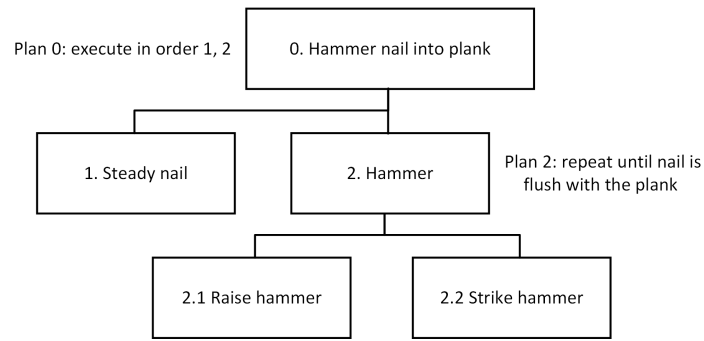

**Fig. 1.** An example of a simple HTA adapted from Stanton [31].

What may reveal itself in the method is some of its difficulty of use [31]. Why not three sub-goals instead of two, for example? And what about the stop rule? How flush should a flush nail be before stopping? But despite its shortcomings, the method's wide-spread use is nevertheless a testament to its undeniable practicality.

Even though being a very popular method in a myriad of different areas [32, 33, [34], HTA has not seen much use in HRI where the robots are social robots. It has seen its use with industry and service robots [35, 36, 37] which is to be expected, since these are more closely related to the kind of technologies usually studied in human factors[1]. Not only HTA, but task analysis in general is fairly rare in the context of social robotics. One of the few occurrences of using explicit task analysis in the design of social robot interactions can be found by Wilson, Tickle-Degnen and Scheutz [38], who used it for understanding a medical sorting task which later was performed by a social robot. The method they used, however, was a free and unstructured one. We believe HTA might in many cases be a better choice for HRI because of its clear instructions and tangible results which can later be used in the design process.

## 3   Role-play and HTA in our project

These two methods, role-play and HTA, has played an integral part in our ongoing research project striving to develop an application, where the robot Furhat acts as an assistant pharmacist and provides medication counselling. This will be showcased by describing our process in simulating a pharmacy in a lab environment followed by our use of HTA to model the behavior of pharmacists.

---

[1] Human factors often being interested in studying in potentially hazardous technologies like airplanes, submarines or nuclear plants make these kind robots more appealing than social ones.

### 3.1   Role-playing an interaction in a pharmacy

The central tenet of human-centered design is designing for humans. To do this in any meaningful fashion you need to understand those you are designing for. Since we wanted to understand the pharmaceutical context, a natural approach would be to video record people, both pharmacist and customers in a real pharmacy, but ethical concerns, such as legislation and customer privacy, make this easier said than done.

In this situation, role-play provided a natural solution to our problem. We opted to set up a simulated environment in our lab where we had 2 actual pharmacists role-play as if they were currently working and served 3 people each[2] who role-played as customers in need of an emergency contraceptive pill (see figure 2).

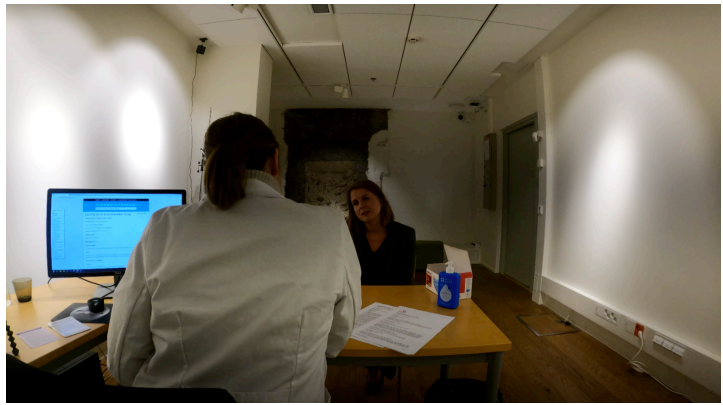

**Fig. 2.** Role-playing the interaction between pharmacist and customer in a lab environment.

The interactions were video recorded from multiple angles, and after the interaction, the customers gave some additional input concerning the experience through interviews.

The use of role-play not only gave us the option of video recording the interaction — it also gave us freedom in adjusting the interaction between pharmacist and customer, for example by varying what type of emergency contraceptive pill the customers needed. It also gave us opportunity to discuss a sensitive topic without psychologically burdening the participants as much as a real interaction would have. Since the situation was a simulated one, they probably also felt they could talk more openly and lose less face than if the interaction was real. As mentioned before, role-play seems especially fitting in sensitive areas like these.

---

[2] The two pharmacists did not serve the same customers.

## 3.2  Hierarchical task analysis of a pharmacist

After collecting the video data, the first author analyzed the video data carefully and followed the recommended steps in constructing an HTA (see e.g., [39] for a guide.) Since the third author is a pharmacist, the model could be further refined as to describe the task well enough from a pharmacist's point of view (see figure 3).

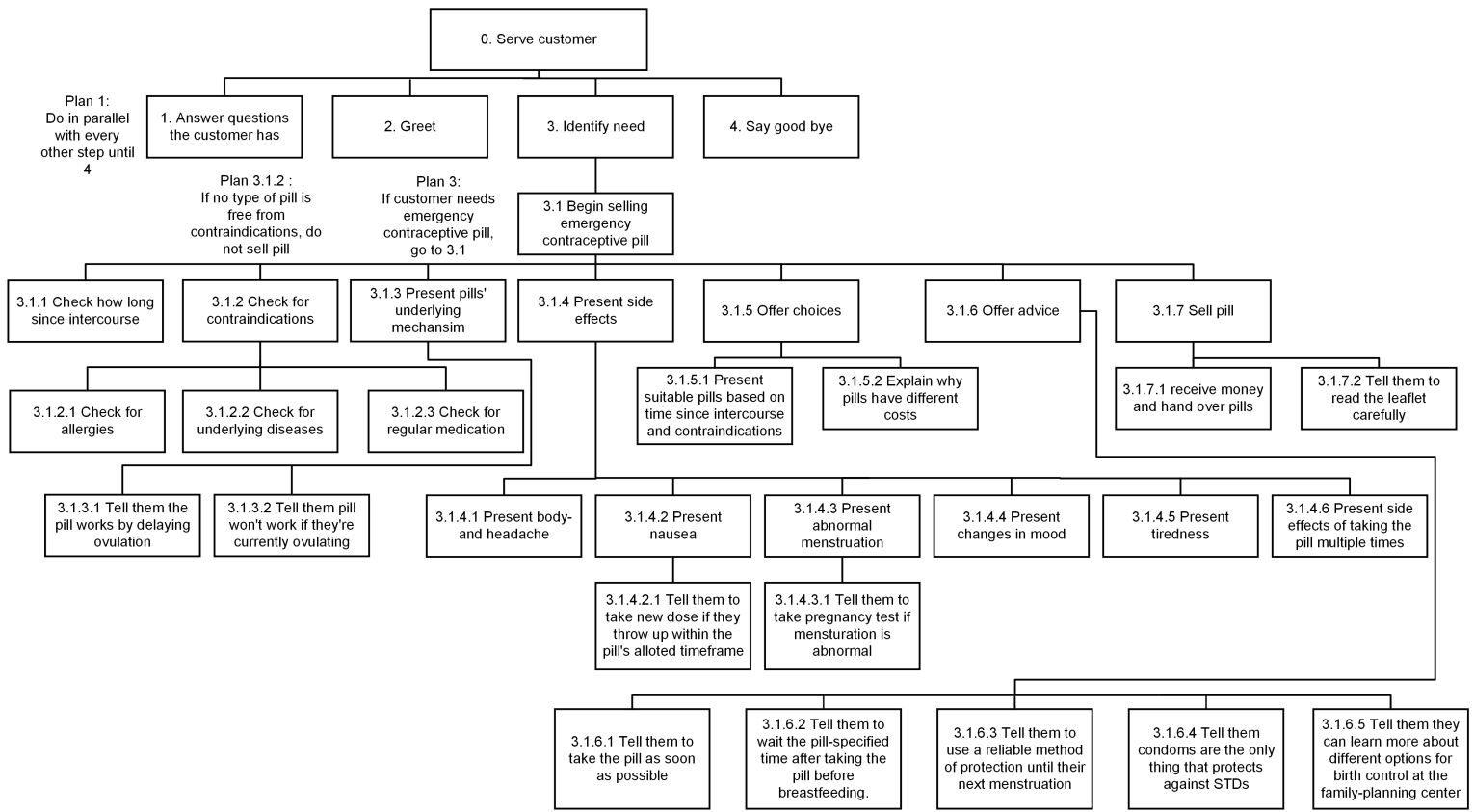

**Fig. 3.** HTA of a pharmacist serving a customer by selling an emergency contraceptive pill. The steps proceed sequentially (*0, 1, 2, 3…*) unless otherwise stated by a plan (*plan 1, plan 3 and plan 3.1.2*).

By constructing an HTA it provided the entire team with an overview of the task and eased the identification of critical points. Its meticulous process is especially suitable for the medical context where potential errors can be detrimental; and since our application is meant to collaborate with pharmacists, it helped us in identifying the situations where a pharmacist needs to take control of the situation. In general, the HTA aided the co-design greatly since it facilitated ideation and collaboration between the team members. We have found HTA so useful that we have developed a method which considers each step in the HTA in greater detail, a method there is not enough room to describe here.

The HTA also served as a clear blueprint for structuring the robots' intended behavior. Since each goal and subgoal in the HTA can be thought of as a state in the robot, this was easily implemented inside the Furhat platform because of its rule- and state-based framework, for example:

```
val Greet = state {
  onEntry {
    furhat.say("Hi, how can I help you?")
    goto(IdentifyNeed)
  }
}
```

In this simple example, the program begins by entering the state *Greet* and greeting the customer (corresponding to sub-goal 2 in the HTA). Furhat then switches to its next state *IdentifyNeed* (3 in the HTA). This approach is not of course limited to rule-based systems but can for example be extended to large language models (LLMs) [40] as well. Another advantage of using HTA is that it can be used in creating Hierarchical Task Networks [41], since this ensures the tasks are properly constructed.

## 4    Discussion and conclusion

We hope that by briefly showcasing our inclusion of role-play and HTA in our design process, we can inspire other researchers to do likewise, especially in the area of social robotics where these methods have not seen much use. The two methods provide methodological validity, as well as practical applicability by ensuring that, firstly, a valid model is used to base the design on, in this case real pharmacists; and, secondly, that the insights gained from observation are implemented and co-design is facilitated through the use of HTA.

In the future, it would be worth examining how role-play could be used further, maybe by developing specific protocols or using it in other sensitive settings. As briefly mentioned before, our team has been developing a method that in greater detail analyses the goals of the HTA. However, what is also needed is task analysis methods which better supports parallel and conditional interaction between multiple actors where later actions depend on previous ones, something which neither HTA nor our method is well-suited for. This need becomes even more apparent if you were to model robots instead of humans[3] and include the interaction between different modules the robots use, for example artificial intelligence models. In cases like these, careful modelling would be needed when describing how the robot coordinates between the use of different modules, like LLMs or action recognition models [42]. HTA does allow some kinds of parallel and conditional goals by implementing plans, but it easily becomes cumbersome and cluttered. Some task analysis methods more suited to this kind of modelling have been developed [43, 44], but they do not allow the practical progression from task analysis to development that we experienced in the case of HTA. It is not trivial to strike a balance between a task analysis method that is neither too abstract so that it loses its immediate practical value, nor too concrete so it becomes overwhelming. Finding the balance, however, would lead to an invaluable tool.

---

[3] We of course do not mean to suggest that robots are more complex than humans, but that modelling humans is by necessity done on a higher level of abstraction than robots.

**Acknowledgments.** This work was funded by the research profile Solutions for Health at Åbo Akademi University [Academy of Finland, project #336355], Svenska kulturfonden and Högskolestiftelsen i Österbotten,

**Disclosure of Interests.** None of the authors have any relevant conflict of interest to declare.